\documentclass[sigplan,screen]{acmart}

\AtBeginDocument{%
  }

\copyrightyear{2025}
\acmYear{2025}
\setcopyright{acmlicensed}
\acmConference[ASPLOS '25]{Proceedings of the 30th ACM International Conference on Architectural Support for Programming Languages and Operating Systems, Volume 1}{March 30--April 3, 2025}{Rotterdam, Netherlands} 
\acmBooktitle{Proceedings of the 30th ACM International Conference on Architectural Support for Programming Languages and Operating Systems, Volume 1 (ASPLOS '25), March 30--April 3, 2025, Rotterdam, Netherlands}
\acmDOI{10.1145/3669940.3707268}
\acmISBN{979-8-4007-0698-1/25/03}

\usepackage[]{hyperref}
\usepackage{multirow}
\usepackage{makecell}
\usepackage{threeparttable}
\usepackage{amsmath}
\usepackage{amsfonts}
\usepackage{algorithm}
\usepackage{algpseudocode}
\usepackage{xcolor}
\begin{document}

\title[Masked Vector Quantization For Efficient DNN Compression and Acceleration]{MVQ: Towards Efficient DNN Compression and Acceleration with Masked Vector Quantization}

\author{Shuaiting Li}
\authornote{Both authors contributed equally to this research.}
\email{list@zju.edu.cn}
\affiliation{%
  \institution{Zhejiang University}
  \city{Hangzhou}
  \state{Zhejiang}
  \country{China}
}

\author{Chengxuan Wang}
\authornotemark[1]
\email{wangchengxuan@zju.edu.cn}
\affiliation{%
  \institution{Zhejiang University}
  \city{Hangzhou}
  \state{Zhejiang}
  \country{China}
}

\author{Juncan Deng}
\email{dengjuncan@zju.edu.cn}
\affiliation{%
  \institution{Zhejiang University}
  \city{Hangzhou}
  \state{Zhejiang}
  \country{China}
}

\author{Zeyu Wang}
\email{wangzeyu2020@zju.edu.cn}
\affiliation{%
  \institution{Zhejiang University}
  \city{Hangzhou}
  \state{Zhejiang}
  \country{China}
}

\author{Zewen Ye}
\email{lucas.zw.ye@zju.edu.cn}
\affiliation{%
  \institution{Zhejiang University}
  \city{Hangzhou}
  \state{Zhejiang}
  \country{China}
}

\author{Zongsheng Wang}
\email{wangzongsheng@zju.edu.cn}
\affiliation{%
  \institution{Zhejiang University}
  \city{Hangzhou}
  \state{Zhejiang}
  \country{China}
}

\author{Haibin Shen}
\email{shen_hb@zju.edu.cn}
\affiliation{%
  \institution{Zhejiang University}
  \city{Hangzhou}
  \state{Zhejiang}
  \country{China}
}

\author{Kejie Huang}
\authornote{Corresponding author.}
\email{huangkejie@zju.edu.cn}
\affiliation{%
  \institution{Zhejiang University}
  \city{Hangzhou}
  \state{Zhejiang}
  \country{China}
}

\renewcommand{\shortauthors}{Shuaiting Li et al.}
\begin{abstract}
Vector quantization(VQ) is a hardware-friendly DNN compression method that can reduce the storage cost and weight-loading datawidth of hardware accelerators. However, conventional VQ techniques lead to significant accuracy loss because the important weights are not well preserved. To tackle this problem, a novel approach called MVQ is proposed, which aims at better approximating important weights with a limited number of codewords. At the algorithm level, our approach removes the less important weights through N:M pruning and then minimizes the vector clustering error between the remaining weights and codewords by the masked k-means algorithm. Only distances between the unpruned weights and the codewords are computed, which are then used to update the codewords. At the architecture level, our accelerator implements vector quantization on an EWS (Enhanced weight stationary) CNN accelerator and proposes a sparse systolic array design to maximize the benefits brought by masked vector quantization.\\
Our algorithm is validated on various models for image classification, object detection, and segmentation tasks. Experimental results demonstrate that MVQ not only outperforms conventional vector quantization methods at comparable compression ratios but also reduces FLOPs. Under ASIC evaluation, our MVQ accelerator boosts energy efficiency by 2.3$\times$ and reduces the size of the systolic array by 55\% when compared with the base EWS accelerator. Compared to the previous sparse accelerators, MVQ achieves 1.73$\times$ higher energy efficiency.  
\end{abstract}



\maketitle
\section{Introduction}
Deep Neural Networks (DNNs) have achieved great success in various applications, such as image classification, object detection, segmentation, etc. Due to the rapid growth of edge intelligent applications, there has been a growing interest in deploying DNNs to edge devices, especially for applications with real-time constraints and privacy issues. 
With the rapid growth of model parameters, the on-chip cache of existing accelerators often cannot store all model parameters and activations~\cite{pullini2019mr, rossi20214}. Consequently, the accelerators need to access off-chip DRAM frequently, resulting in increased latency and power consumption. 

To achieve low-power computing, it is generally necessary to begin with model optimization and develop hardware-friendly compression algorithms. However, existing model compression methods face challenges in balancing compression rates and accuracy. For instance, 1-bit quantization reduces the model size by 32$\times$, but low-bit quantization can cause significant quantization errors, leading to a significant drop in accuracy. 
Unstructured pruning~\cite{han2015deep,louizos2017learning,renda2020comparing} has to store the indices of unpruned weights while the pruning rate of structured pruning~\cite{li2016pruning, molchanov2019importance} rarely exceeds 50\%.
Vector Quantization (VQ), by grouping and clustering weights, has the potential to achieve higher compression ratios and is a hardware-friendly compression method. However, existing vector quantization works ~\cite{gong2014compressing,wu2016quantized,Martinez_2021_CVPR,son2018clustering,stock2019and, chen2020towards} treat all weights equally. As a result, many important weights are forced to align with unimportant weights, leading to numerous clustering errors and significant accuracy degradation. 
 
To achieve extreme compression of neural network models while keeping high accuracy and hardware-friendliness, a Masked Vector Quantization (MVQ) method is proposed. Our method aims to better approximate the important weights in DNNs. The first step in the codebook generation process is the N:M pruning, where unimportant weights are removed. The masked k-means algorithm is then applied to prevent unimportant weights from interfering with the vector clustering process. Only unpruned weights in subvectors need to be computed with codewords. The codewords are then updated. Quantization is applied to the codebook to ensure hardware-friendliness. Codewords of the codebook can be fine-tuned by masked gradients from corresponding sub-vectors of compressed weights.

To fully leverage the potential of our MVQ compression method, we propose a SW-HW co-designed accelerator based on the EWS dataflow~\cite{wang2024ews}. This architecture incorporates an assignment-aware weight-loading controller to minimize the DRAM data access cost, along with a sparsity-aware systolic array that significantly reduces computing resources. This architecture facilitates achieving high area and energy efficiency. 

Experimental results indicate that when our MVQ algorithm is applied to various models, it not only improves accuracy but also reduces FLOPs compared to conventional VQ methods. Our accelerator is synthesized using a 40 nm 0.99V library. In comparison to the base EWS accelerator, our MVQ accelerator enhances energy efficiency by 2.3 $\times$ and reduces the size of the systolic array by 55\%. Furthermore, our accelerator outperforms previous sparse accelerators, achieving over 1.73 $\times$ higher efficiency.

The main contributions of the paper are summarized as follows:
\begin{itemize}
  \item We introduce a novel compression pipeline and the masked vector quantization algorithm to better approximate important weights during vector clustering.
  \item We propose an area- and energy-efficient hardware accelerator for this MVQ compression algorithm. 
  \item We validate the accuracy improvement of our algorithm as well as the efficiency gains of our accelerator.
\end{itemize}

\section{Related Work}
In this section, we first provide a brief review of popular compression techniques for convolutional neural networks. Then, we review related ASIC accelerators and Hardware-Software (HW-SW) co-design approaches for efficient inference.

\subsection{N:M Pruning.} 
The fine-grained N:M pruning technique performs unstructured pruning within each group while enforcing an additional constraint that only N values are pruned. N:M sparsity can be accelerated on modern hardware ( Nvidia Ampere Sparse Tensor Core~\cite{mishra2021accelerating}) while maintaining high sparsity. ~\cite{zhou2021learning} trains the N:M sparse model with uniform sparsity from scratch. ~\cite{sun2021dominosearch} finds mixed layerwise N:M sparsity patterns to achieve higher accuracy and compression ratio.

\subsection{Vector quantization.}
Instead of storing weights themselves, vector quantization stores a codebook and a list of assignments. The compressed weight is reconstructed by looking up the codeword in the codebook that corresponds with the assignment. BGD~\cite{stock2019and} uses the product quantization \cite{jegou2010product} method and adopts activation into clustering to minimize reconstruction error. Additionally, layer-wise fine-tuning of codewords via distillation is applied to ensure that the output of each layer is similar to that of the pre-trained model. PQF ~\cite{Martinez_2021_CVPR} focuses on identifying parameter groups to compress together, searching for optimal permutations to enable efficient compression. DKM~\cite{cho2022dkm} casts k-means clustering as an attention problem and enables joint optimization of the DNN parameters and clustering centroids. However, in conventional vector quantization, a problem arises when important and unimportant weights align within subvectors, causing significant errors and decreased accuracy.

\subsection{Systolic Array-based accelerators for CNNs}
Designing dedicated dataflows to fully utilize the data reusability in CNN for memory access optimization is a significant research topic in using systolic array-based CNN accelerators. TPU~\cite{jouppi2017datacenter} is a WS dataflow accelerator proposed by Google. It parallelly unfolds in the input channel(C) and output channel(K) of the convolution kernel, also known as the C|K unfolding. Gemmini~\cite{gemmini} further reduces the number of pipeline registers and pipeline latency in the systolic array by introducing adjacent PE to form a combinational logic addition tree of tiles. EWS~\cite{wang2024ews} observes the problem of frequent array accesses to global buffers caused by insufficient data reusability in WS data flow CNN accelerators with C|K unfolding. It proposes an Enhanced Weight Stationary (EWS) that fuses IS and OS strategies.

\subsection{Accelerators with compression-based co-design}
Another hot topic is HW-SW co-design accelerators based on various model compression techniques.  
~\cite{liu2020systolic} proposed a Systolic Tensor Array (STA), based on the N:M pruning, operating with OS data flow. When using 2:4 pruning, only 2 multipliers and 1 adder are needed to implement 4×4 vector dot product operations, reducing the computational resources by about half compared to non-sparse models.~\cite{liu2022s2ta} further exploit Density Bound Block (DBB) sparsity for both weights and activations. ~\cite{im2023sibia} explores efficient signed bit-slice architecture with signed bit-slice representation (SBR) for efficient dense DNN acceleration. ~\cite {lee2019convolutional, wu2019accelerator} design vector quantization-based accelerators, but their vector quantization algorithms suffer significant accuracy degradation, and the implemented accelerators do not optimize the dataflow specifically based on the parameter characteristics of CNNs.

\section{Preliminaries}
For a typical vector quantization algorithm, $B\in R^{NG\times d}$ denotes a 2D data matrix, $C = \{c_1,...,c_k\}$ denotes the codebook containing $k$ codewords of size $d$. We can view $B$ as a collection of $NG$ subvectors of dimension $d$. Each subvector $b_j \in B$ is mapped to one of the codewords $c_i \in C$, the function can be represented as $q(b_j) = c_i$. In the subsequent sections of this article, we will consistently represent the number of codewords as $k$, the block size (length of a codeword) as $d$, and the number of subvectors as $NG$. 

To minimize the Sum of Square errors (SSE), the most popular approach is the k-means clustering algorithm. The basic procedure of k-means is as below:
\begin{enumerate}
    \item Randomly select $k$ subvectors as initial codewords.
    \item For each subvector, calculate the Euclidean distance between itself and every codeword, then assign itself to the nearest codeword.
    \item For each codeword, calculate the average value of subvectors assigned to it.
    \item Repeat step2 and step3 until the number of changing assignments falls below a given threshold, which is typically set to 0.1\% of the total number of subvectors.
\end{enumerate}

\section{Masked Vector Quantization}
In this section, we first conduct empirical experiments to reveal reducing clustering errors of important weights is crucial to model performance. Then, we present our masked vector quantization pipeline in detail.
\label{section_algorithm}

\subsection{Empirical observation}
\begin{figure}[h]
    \centering
    \includegraphics[width=0.95\linewidth]{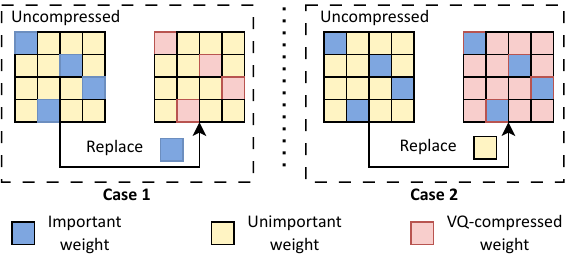}
    \caption{Two cases to replace some weights with corresponding vector-quantized ones}
    \label{replace}
\end{figure}
\begin{table}[h]
    \centering
    \small
    \renewcommand{\arraystretch}{0.6}
    \setlength{\tabcolsep}{3mm}
    \begin{tabular}{c|cc|cc}
    \toprule
    \multirow{2}{*}{Methods} & \multicolumn{2}{c|}{ResNet-18} & \multicolumn{2}{c}{ResNet-50}\\
    \cmidrule{2-5}
    & SSE &  Acc & SSE & Acc\\
    \midrule
    Case 1 & 576 & 5.8 & 695 & 1.26\\
    \midrule
    Case 2 & 623 & \textbf{37.46} & 771 & \textbf{55.39}\\
    \bottomrule
    \end{tabular}
    \caption{Partly vector-quantized accuracy on ImageNet dataset across two cases and different models. Note that we didn't perform any fine-tuning in this experiment.}
    \label{emprecial_experiment}
\end{table}
To investigate the influence of clustering errors on different weights, we conduct preliminary experiments on the ImageNet dataset. We present a simple visualization in Fig.~\ref{replace} illustrating two scenarios of introducing clustering errors to weights of varying significance. We select 2 weights with the highest magnitude within each of the 8 continuous elements. These selected weights, which account for 25\% of total weights, are marked as important weights, while the others are marked as unimportant weights. We perform a layerwise vector quantization for convolutional layers with $k=512$ and $d=8$ and obtain vector-quantized weights. We replace part of the weights with their corresponding quantized weights. In Case 1, important weights are replaced with their vector-quantized weights, while in Case 2, the remaining unimportant weights are replaced with their vector-quantized weights. Without any fine-tuning, we evaluate the top-1 accuracy and SSE and summarize the results in Tab.~\ref{emprecial_experiment}. Even though case 2 exhibits higher SSE, it achieves significantly higher accuracy than that of case 1. Therefore, we can conclude that approximating important weights during vector quantization plays a dominant role in the preservation of model performance. Simply reducing overall clustering error leads to a suboptimal solution.

\subsection{Overview of MVQ}

\begin{figure}[h]
    \centering
    \includegraphics[width=0.95\linewidth]{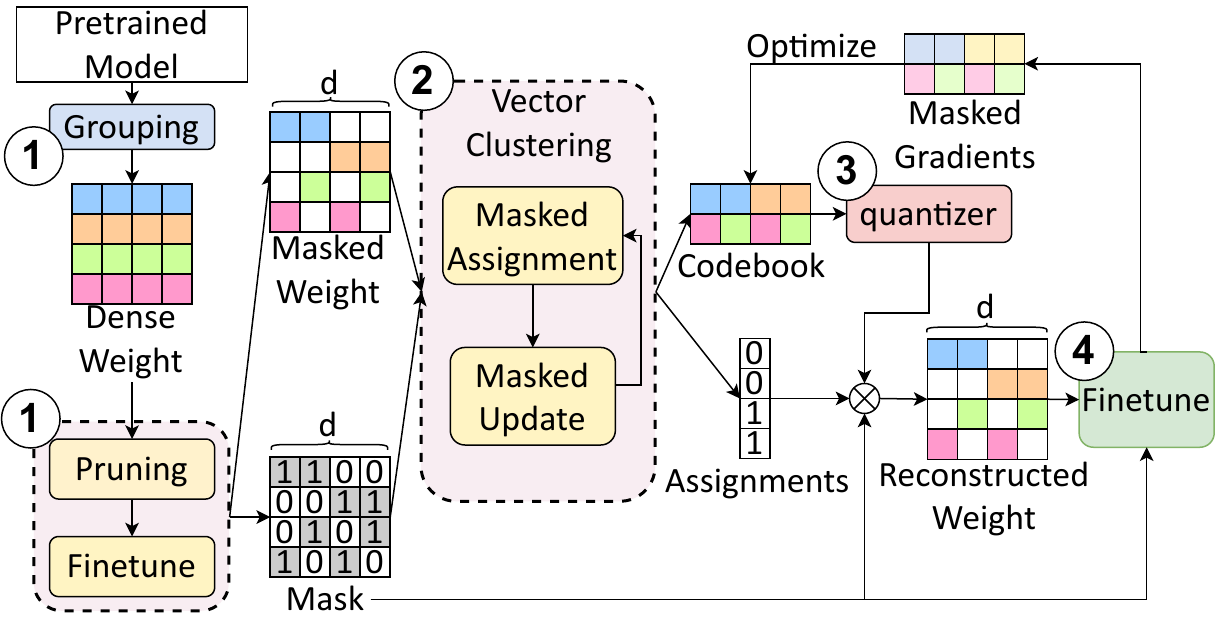}
    \caption{Overall compression pipeline of our MVQ algorithm. 1. The grouping strategy is determined for both pruning and vector clustering. Pruning is performed and the sparse model is fine-tuned. 2. Unpruned weights and the mask are incorporated into the masked k-means algorithm to generate the codebook. 3. Symmetric 8-bit quantization is applied to the codebook to ensure hardware-friendliness. 4. The codebook is fine-tuned with masked gradients.}
    \label{picture:whole pipeline}
\end{figure}

Based on the conclusion from the previous subsection, our method prioritizes minimizing errors for important weights. To accomplish this, we propose MVQ, which uses pruning to remove unimportant weights. Then, pruning masks are used in the masked k-means process to avoid aligning important weights with unimportant ones. 
As illustrated in Fig.~\ref{picture:whole pipeline}, our compression pipeline consists of four steps: 
 
1. Weight grouping and pruning: The codebook's encoding is established and weights are divided into groups. Subsequently, less important weights are removed through pruning.
 
2. Masked k-means clustering: To avoid the influence of these pruned weights, the masked k-means algorithm is applied to execute vector clustering. 
 
3. Codebook quantization: Symmetric 8-bit quantization is applied to the codebook to ensure hardware-friendliness.
 
4. Finutuning: To recover the accuracy of the compressed model, codewords are fine-tuned using masked gradients.
 
Detailed explanations of these steps will be provided in the following subsections.

\subsection{Weight grouping and Pruning}
\begin{figure}[h]
    \centering
    \includegraphics[width=0.85\linewidth]{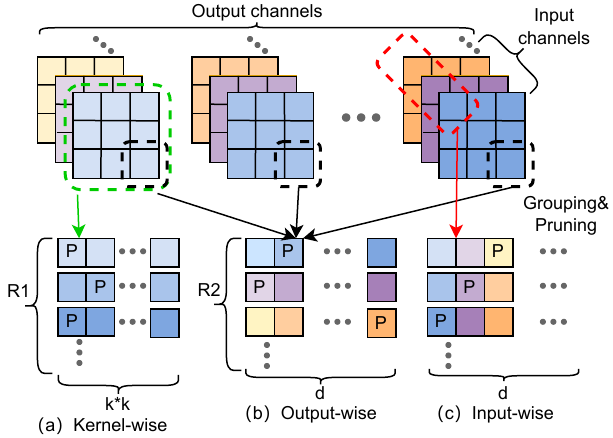}
    \caption{An illustration of grouping strategies. For kernel-wise grouping, $R_{1}=C_{out}\times C_{in}$. For input and output-channel-wise grouping, $C_{out}$ and $C_{in}$ are multiples of $d$, and $R_{2}=\frac{C_{out}}{d} \times {C_{in}} \times {k} \times k$.}
    \label{grouping}
\end{figure}

To perform subsequent vector clustering, weight $W$ has to be firstly grouped to a set of subvectors with length $d$. As illustrated in Fig.~\ref{grouping}, the 4D weight can be split and grouped into a set of vectors in different dimensions, including kernel dimensions, input-channel dimensions, and output-channel dimensions. The group length $d$ is constrained to $k*k$ for kernel dimensions. However, the group length of both output-channel-wise and input-channel-wise grouping is more flexible, i.e., it could be a power of 2. In pursuit of a higher compression ratio and hardware efficiency, the channel-wise grouping strategy is chosen. In this way, the 4D weight is reshaped into a 2D matrix $W_{r}$ of size $(\frac{Cout}{d}\times C_{in} \times k \times k) \times d$. 

After grouping, N:M pruning, which achieves a high level of sparsity while maintaining a regular sparse structure, is applied. For each subvector in $W_{r}$ of length $d$, $\frac{(M-N)\times d}{M}$ elements are pruned, and $d$ must be multiples of $M$. The sparse model is then fine-tuned before undergoing the vector clustering process. Additionally, a bitmask denoted as $BM$ is saved to indicate the positions of the pruned and unpruned weights. The same grouping strategy applied to weights reshapes $M$ to $M_{r}$. The 2D weight matrix after pruning and fine-tuning is denoted as $W_{rp}$. Each subvector $w_j$ from $W_{rp}$ is assigned to one of the codewords, namely $q(w_j) = c_i$, where $i$ is the assignment. The corresponding mask of $w_j$ is denoted as $bm_j$. The detailed settings of pruning strategies and fine-tuning strategies will be discussed in Section 6.2.

\subsection{Masked k-means clustering}

\begin{figure}[!h]
    \centering
    \includegraphics[width=0.95\linewidth]{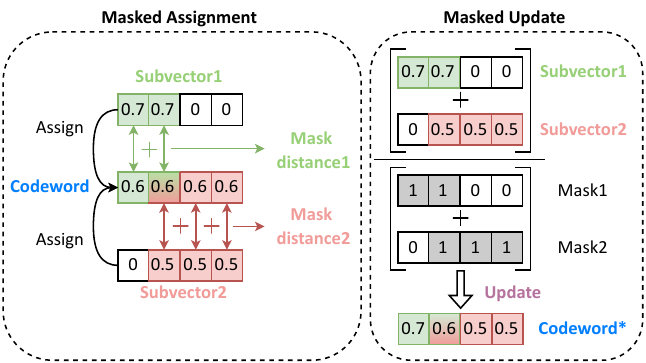}
    \caption{A simple example of masked k-means. For subvector1 and subvector2 assigned to the same codeword, green blocks represent unpruned weights in subvector1, while red blocks represent unpruned weights in subvector2. During the masked assignment step, only the unpruned weights in subvectors are used to calculate distances with the codeword. 
    During the masked update step, only unpruned weights are involved in updating the codeword. }
    \label{picture:mask_kmeans}
\end{figure}

Upon obtaining the sparse model and its corresponding mask, our objective is to minimize the clustering error for important weights by minimizing the following objective function:
\begin{equation}
\|W_{rp}-\widehat{W_{rp}}\|_2^2 = \sum_j\|w_j-q(w_j)\circ {bm_j}\|_2^2
\end{equation}
Therefore, the codebook and assignments can be learned using the k-means algorithm. However, if common k-means clustering (as discussed in preliminaries) is applied to $W_{rp}$, many important weights have to be aligned to zero, making it difficult to reduce the clustering error for these weights (this phenomenon will be further demonstrated in the ablation study section). To reduce the impact of pruned weights, our masked k-means algorithm replaces these two steps with the masked assignment and the masked update:

\textbf{Masked assignment.}
Each subvector $w_j$ is assigned to the nearest codeword $c_i$ such that
\begin{equation}
    c_i = \arg\min_{c\in C} \|w_j - c\circ{bm_j}\|_2^2 \label{equation:masked assignment}
\end{equation}

As shown in Fig.~\ref{picture:mask_kmeans}, only the unpruned weights in $w_j$ are considered when calculating the Euclidean distance from weight subvector to each codeword. This step can be computationally intensive as the codebook needs to be multiplied with different masks for each subvector. To efficiently implement this step on GPU, we use the broadcasting mechanism in PyTorch. we construct a tensor of shape $\lbrack L, k, d \rbrack$ to depict the application of different masks on each codeword. Then, we employ $torch.cdist$ function to compute its Euclidean distance with the weight matrix. Additionally, we use batch processing to reduce memory consumption.

\textbf{Masked update.}
Let $v_p$ denote the subvectors that are currently assigned to $c_i$. Let $n_p$ denote the mask for $v_p$. The update of the codeword is a least square problem, and its analytical solution is the average of corresponding subvectors. 
\begin{equation}
    c_i^* = \arg\min_{c_i\in R^d}\sum_p\|(v_p - c_i)\circ n_p\|_2^2 \label{equation:MSE solution}
\end{equation}
The pruned weights don't rely on the codeword for reconstruction because they are permanently zero. Therefore, in the masked update step, the condition in the least squares problem is changed to Eq.~\ref{equation:MSE solution}. The analytical solution to Eq.~\ref{equation:MSE solution} is computed as follows:
\begin{equation}
    c_i^* = \frac{\sum_p v_p}{\sum_p n_p} \label{equation:masked average}
\end{equation}

As shown in Fig.~\ref{picture:mask_kmeans}, different from simply averaging $v_p$, only the average of the unpruned element is calculated, which prevents the great amount of 0 values from disturbing $c_i^*$. Division in Eq.~\ref{equation:masked average} refers to element-wise division. Once the vector clustering is done, the assignments and the bitmask are fixed. Detailed vector clustering error comparison will be provided in the ablation study section.

\begin{figure}[]
    \centering
    \includegraphics[width=0.95\linewidth]{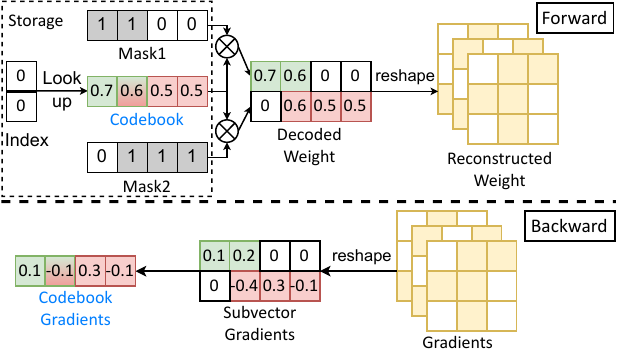}
    \caption{A simple example of the fine-tuning process. Weights are reconstructed from the codebook, assignments, and masks for forward computation, while masked gradients are calculated to update the codebook.}
    \label{picture: reconstruct}
\end{figure}

\subsection{Codebook quantization}
Quantization can be further applied to the codebook. Given the full-precision data $v$, the scaling factor $s$, and the quantization bits $qb$, the symmetric uniform quantized representation $\hat{v}$ is calculated as :
\begin{equation}
    \hat{v} = s_{w}\times clamp(\lbrack\frac{v}{s_{w}}\rbrack;-2^{qb-1},2^{qb-1}-1])
    \label{equation: codebook quantization}
\end{equation}
 To ensure hardware compatibility, signed 8-bit quantization as illustrated in Eq.~\ref{equation: codebook quantization} is applied. One codebook share the same $s_{w}$, and LSQ~\cite{esser2019learned} is used to determine $s_{w}$. For large values of k, the storage cost of the codebook itself becomes an unignorable factor. Quantizing the codebook also helps to reduce this cost. 

\begin{figure*}[h]
    \centering
    \includegraphics[width=0.98\textwidth]{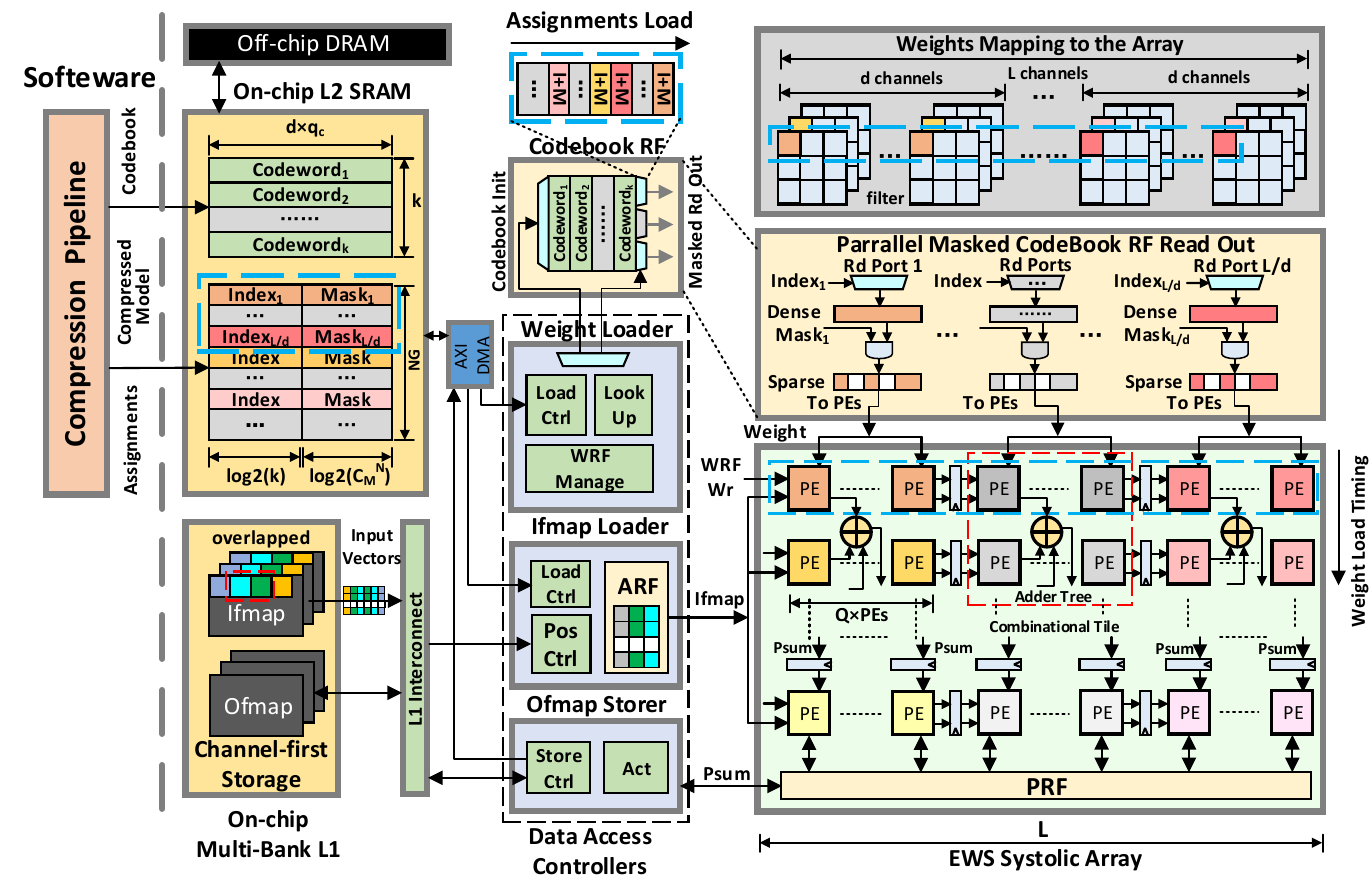}
    \caption{Overview of our efficient accelerator with masked vector quantization.}
    \label{accelerator_overview}
\end{figure*}

\subsection{Fine-tuning method}
To restore the network's original accuracy, it is crucial to fine-tune the codebook. Our method’s final storage comprises three components: assignments, codebooks, and masks. During the forward pass, as depicted in Fig.~\ref{picture: reconstruct}, codewords retrieved from the codebook perform a bit selection operation with their corresponding mask to obtain decoded weights. These decoded weights can be reshaped back to 4D reconstructed weights, which are then used in forward computation. 

During the backward propagation, let $Loss$ be the loss function of the network. Note that $Loss$ is differentiable to each of the learned centroids. As shown in Fig.~\ref{picture: reconstruct}, to avoid the distribution of zero gradients of pruned weights, masked gradients are computed for each codeword, in a similar form to Eq.~\ref{equation:masked average}. Then, gradients-based learning is performed as follows :
\begin{equation}
    c_i \gets c_i- O \left( \frac{\sum_p \frac{\delta Loss}{\delta v_p} \circ n_p}{\sum_p n_p}, \theta \right)
\end{equation}
where $O(,)$ is the optimizer ( Adam~\cite{kingma2014adam}, SGD, AdamW) with hyperparameter $\theta$ (learning rate, momentum, weight decay).

\section{Accelerator Implementation for MVQ}
In this section, we display our hardware microarchitecture design and showcase the hardware-software co-design involving masked vector quantization. 
Denote $b_f$ as the bit-width of full-precision weight. Given a weight block and a mask block of the same size $NG\times d$. The Compression Ratio of our method is calculated as:
\begin{equation}
    Comp. Ratio = \frac{NG*d*b_f}{b_a + b_m + b_c} 
\end{equation}
where $b_a$, $b_m$, and $b_c$ are the storage costs of assignments, masks, and codebooks, respectively. Assuming that the weight block is clustered into $k$ codewords and the codebook is quantized to $q_c$ bits, we can calculate $b_a = \lceil\log_{2}(k)\rceil\times NG$ and $b_c = k\times d \times q_c$. 
Note that $b_c$ is relatively small, quantizing the codebook noticeably improves the compression ratio only when $k$ is very large. 
As for $b_m$, storing the complete bitmask requires one bit per weight, which is unbearable for extreme compression. Since a N:M pruned block contains only $C_{M}^{N}$ combinations of masks, a look-up table could be created, reducing the storage cost to $\frac{\lceil\log_{2}C_{M}^{N}\rceil}{M}$ bit per weight.
\subsection{Arthitecture Overview}
After obtaining the compressed model through the method in Section~\ref{section_algorithm}, we load the two parts of the model: Codebook and Assignments (including the index and mask required to reconstruct each weight vector of length $d$) from the off-chip memory to the on-chip L2 SRAM. An overview of our accelerator is shown in Fig.~\ref{accelerator_overview}. The accelerator consists of data access controllers (Weight/Ifmap Loader and Ofmap Storer) that support EWS dataflow, a systolic array with combinational tile referring to Gemmini~\cite{gemmini} and PE with Weight Register Files (WRF). The data access controllers implement data interaction between L2 and L1 and the array through DMA 
with a datawidth of 64-bit. 
L1 is a global buffer composed of multiple SRAM banks, providing the accelerator with enough datawidth to read Ifmap and store Ofmap/Psum with a single-cycle delay. 

\begin{figure}[h]
    \centering
    \includegraphics[width=0.9\linewidth]{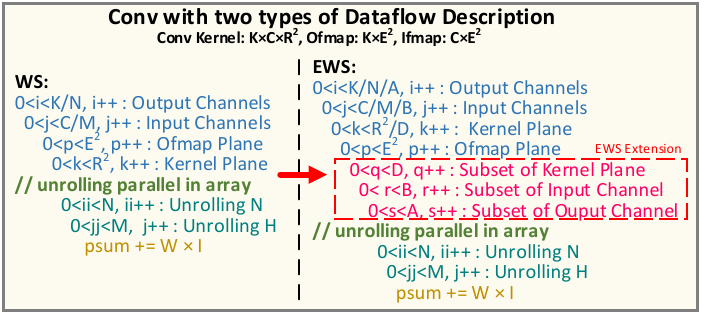}
    \caption{Brief description of EWS dataflow}
    \label{dataflow}
\end{figure}
The EWS dataflow deployed by the accelerator refers to the implementation in ~\cite{wang2024ews}, as described in Fig.~\ref{dataflow}. Given the parameters of a convolution layer, compared with the conventional WS, EWS introduces a three-dimensional layer-wise optimizable extension in the red box, which can further split the three dimensions of output channel, input channel and kernel plane by integrating them into the inner loop of feature map plane traversal: In every consecutive $A$ cycles, the activations fetched by systolic array stays in the PEs, so that the PEs can switch weights from the WRFs inside them to multiply with the same activations and get psums corresponding to $N\times A$ output channels. Then in every consecutive $A \times B$ cycles, weights corresponding to $B$ input channels are switched, the accumulated psums can be read or written in the PRFs instead of on-chip SRAM. Finally, $D$ coordinates on the convolutional kernel plane is traversed in consecutive $A \times B \times D$ cycles. As a result, during the traversal of this subset with $A\times B \times D$ weights, the frequency of accessing activation and partial sum from on-chip SRAM can be reduced to the original $1/(A\times D)$ and $1/(B\times D)$ respectively. The extensions $A$ and $B$ can make the PE array with the physical size of $H\times L$ have data reuse the same as that of $(H\times B)\times (L\times A)$ array. The extension $D$ supports the WRF in storing multi-element convolution kernel plane subsets. The Activation Register File (ARF) and Partial Sum Register File (PRF) in Fig.~\ref{accelerator_overview} provide architectural support for the convolutional reuse of Ifmap and channel-wise partial sum accumulation, respectively. To handle the potential splitting and merging of feature maps in the channel, width, and height dimensions during convolution calculation, a specialized DMA is employed, capable of directly transferring 3D tensors between L2 and L1. The controller and load/store components can initiate transfer requests to the DMA through the DMA interconnect, which can respectively handle state queries and data access instructions.

\subsection{Assignment-aware Weight Loading}
To harvest gains for efficient inference from the MVQ compressed model, we have developed a set of methods to map the compression model to the accelerator.

Before the inference starts, the Weight Loader initializes the codebook to the Codebook Register File (CRF) through DMA. Considering that the size of the codebook is usually only at the KB level, and the codebook only needs to be updated once when a layer or even the entire network is executed, the delay and energy overhead caused by initializing the CRF is negligible. The MVQ model requires the assignment of reconstructing a weight vector with a dimension of $d$. Therefore, we integrate the reconstruction of weight vectors into the loading assignment process. Weight Loader reads assignments from L2. The index is used as the address to read CRF, and the mask is stored in $log2(C_M^N)$ bits. After restoration through the Look Up Table (LUT) in the Weight Loader, the sparse mask with $d$ bits is obtained. The sparse reconstructed weight is obtained through the AND gates.

Considering that the width $L$ of the systolic array (corresponding to the parallelism on the convolution output channel) may not be the same as the length $d$ of the weight vector in the MVQ method, a row of PEs requires $L/d$ weight vectors, so the CRF is designed with $L/d$ read ports. In addition, the storage format of the assignments stored in L2 is aligned with the hardware array width $L$, as shown in the blue box in Fig \ref{accelerator_overview}: Each color block represents a d-dimensional weight vector corresponding to $d$ output channels.
\subsection{Sparsity-aware Systolic Array Design}

\begin{figure}[h]
    \centering
    \includegraphics[width=0.8\linewidth]{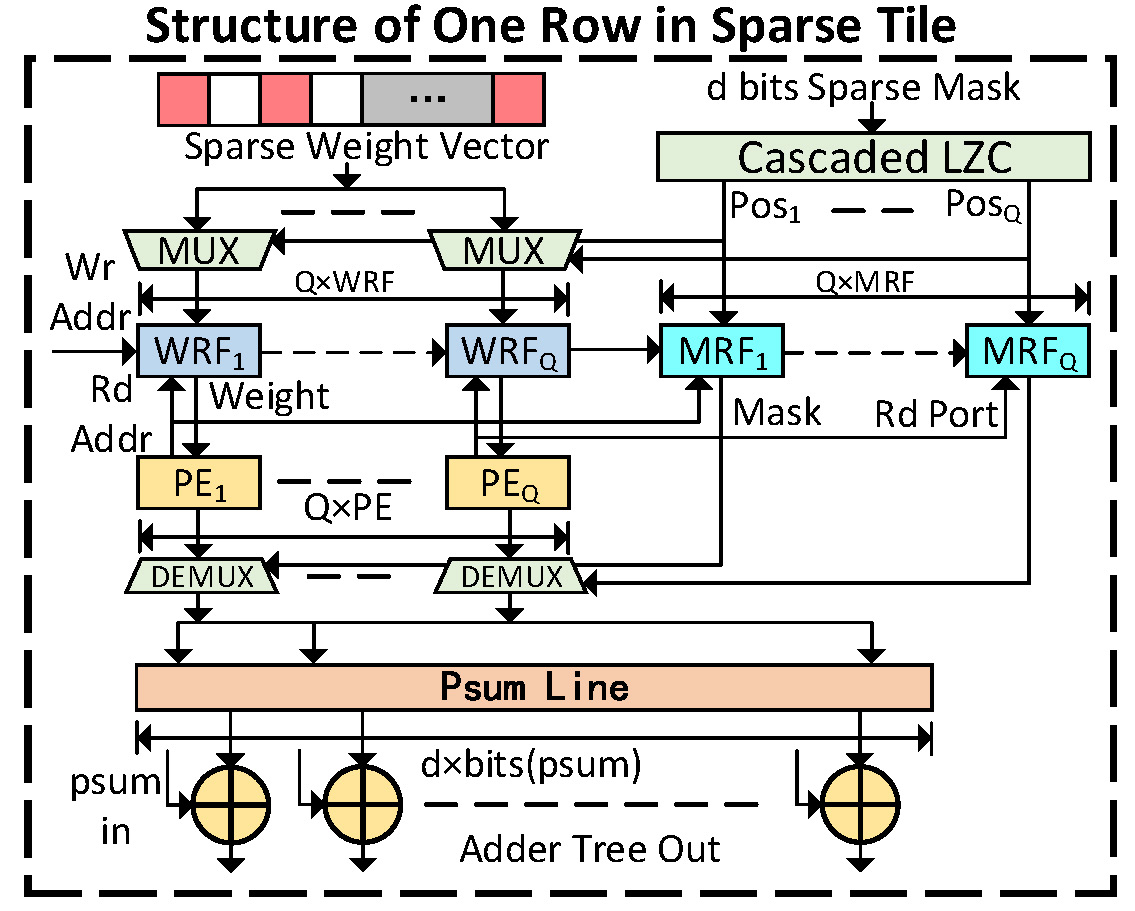}
    \caption{Implementation of the sparse tile.}
    \label{sparsetile}
\end{figure}

Every weight subvector written into the systolic array employs N:M sparsity, however, the computation of 0 elements is meaningless. For EWS dataflow, since the weights in the systolic array are stationary, only $Q=\frac{N}{M} \times d$ numbers of PEs are running in a PE group corresponding to a subvector. Theoretically, we only require $\frac{N}{M}$ of the original computing sources, which means $Q$ PEs for every $d$ output channel.

Every N:M sparsity mask contains N valid bits and thus can't encoded with a common one-hot encoder. We employ $Q$ numbers of LZC (Leading Zero Counter) to encode the sparsity mask with $Q$ valid bits. Each output encoding of LZC is converted into one-hot encoding and sent to the next level LZC with the XOR logic of the input mask at the current level. Q LZCs are cascaded to form the encoder of the sparsity mask. 
\begin{table}[h]
\small
    \centering
    \renewcommand{\arraystretch}{0.6}
    \begin{tabular}{c|c|c}
    \toprule
    Methods & EWS & EWS-Sparse\\
    \midrule
    Multiplier & $H\times d $ & $H\times Q$\\
    \midrule
    Adder & $H\times d$ & $H\times d$ \\
    \midrule
    RF & $H\times d\times 16\times b_{w}$ & \makecell{$H\times Q\times 16\times b_{w}$ +\\ $H\times Q\times 16\times \log_{2}{d}$}\\
    \midrule
    LZC & $NA$ & $H\times Q$ \\
    \midrule
    DEMUX & $NA$ & $H\times Q \times b_{psum}$ \\
    \midrule
    MUX & $NA$ & $H\times Q \times b_{w}$ \\
    \midrule
    Parallelism & $2\times H \times d$ & $2\times H \times d$ \\
    \bottomrule
    
    \end{tabular}
    \caption{Resources Comparsion for a H$\times$d Tile}
    \label{tile resources summary}
\end{table}

The implementation of the sparse tile is illustrated in Fig.\ref{sparsetile}. For a simple explanation, we separate the registers in the PE to the outside of the PE. Based on $Q$ PEs and $Q$ Weight Register Files (WRFs), $Q$ Mask Register Files (MRFs) with a storage bit width of $\log_{2}{d}$ are introduced. The cascaded LZC encoder receives d-bit sparsity masks dispatched to the tile by the weight loading controller as input, and $Q$ position encodings are output and written to the MRF. The MRF and WRF are written and read simultaneously, which simplifies the implementation of the control logic: MRF and WRF share the write address and write enable signals from the array controller, and also share the read address signals from the PE. The output of the multiplier in the PE needs to be coordinated with $Q$ DEMUXs and the MRF output sparsity position encoding that shares the read address with the WRF. These $Q$ partial sums are then distributed to the corresponding partial sum positions as the input for adder trees with a depth of $d$.

\begin{figure}
    \centering
    \includegraphics[width=0.8\linewidth]{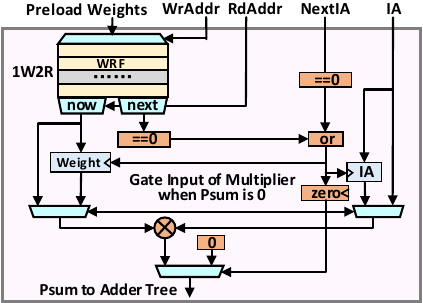}
    \caption{Implementation of the PE with zero value gating.}
    \label{PE_specific}
\end{figure}

The resources cost of an H$\times$d tile is summarized in Tab.~\ref{tile resources summary}. Compared with EWS, our EWS-Sparse greatly reduces the cost of multiplier and register files while maintaining the same computability with sparse models.

To further reduce energy consumption, a zero-value gated PE is implemented to reduce the flip rate of MAC. Quantized weights and input activations naturally exhibit sparsity, if either one is zero in the same computation cycle, the partial sum for this computation will also be zero. The implementation is illustrated in Fig.\ref{PE_specific}. The WRF of the PE is designed with two sets of read ports: one for reading the weights required for the current computation, and another for prefetching the next weight value. This prefetching allows for the determination of whether the subsequent weight value is zero. A similar approach is applied to the input features. When the partial sum of the next cycle is zero, the weight and input stored in registers will be selected, and the input value of the multiplexer at the next cycle will remain unaltered. Consequently, the switching power consumption of multiplexers can be reduced.

\section{Performance Analysis for masked vector quantization Algorithm}
\subsection{Experiment setup}
In our experiments, the accuracy and storage cost are evaluated as key metrics. We commence our experiments by discussing the appropriate pruning strategy. We then perform a detailed ablation study to assess the effect of our masked vector quantization method. Once the optimal compression settings are obtained, we compare the accuracy and FLOPs between our method and other advanced compression techniques on a large set of models.

\begin{figure}[h]
    \centering
    \includegraphics[width=0.95\linewidth]{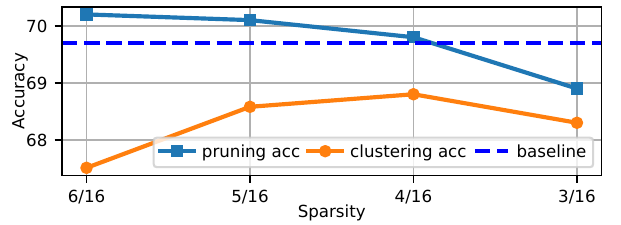}
    \caption{Pruning strategy experiments on ResNet-18.}
    \label{4_16pruning}
\end{figure}
\begin{figure}[h]
    \centering
    \includegraphics[width=0.95\linewidth]{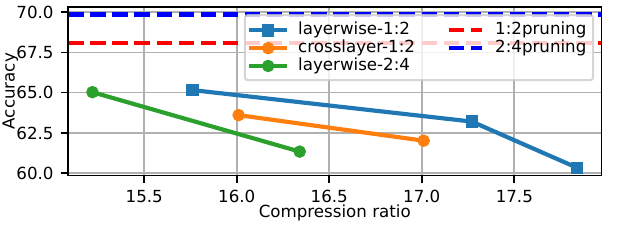}
    \caption{Pruning strategy experiments on MobileNet-v2.}
    \label{1_2pruning}
\end{figure}
\subsection{Pruning strategy}
When the pruning rate is higher, the accuracy of pruning decreases, while the clustering error becomes smaller, narrowing the gap between clustering accuracy and pruning accuracy. Also, different pruning patterns affect the storage cost of masks. Therefore, in our work, selecting a pruning strategy is complex and involves multiple trade-offs.
For classification tasks, we train the sparse model using SR-STE, following the training settings in ~\cite{sun2021dominosearch}. For detection and segmentation tasks, we found that SR-STE is unstable and less effective. Therefore, we simply used the ASP method. 

For models with high redundancy (e.g. ResNets), we seek to achieve highest possible pruning rate while maintaining accuracy, as a higher pruning rate leads to smaller clustering errors. As illustrated in Fig.~\ref{4_16pruning}, pruning rates spanning from 6:16 to 3:16 were tested on ResNet-18. The pruning accuracy rapidly drops below the baseline after sparsity exceeds 75\%. Ultimately, 4:16 pruning yields best clustering accuracy.

For parameter-efficient models (e.g. MobileNets and EfficientNets), 50\% sparsity already results in a notable decline in pruning accuracy. We conducted experiments on MobileNet-v2, exploring both 1:2 and 2:4 pruning. The 2:4 pruning approach demonstrated a 1.7\% higher pruning accuracy but came with 0.25bit/w additional mask storage cost compared to 1:2 pruning. Our findings suggest that 1:2 pruning may offer a more optimal balance between storage and accuracy.  

\subsection{Ablation study}
\begin{figure}
    \centering
    \includegraphics[width=0.9\linewidth]{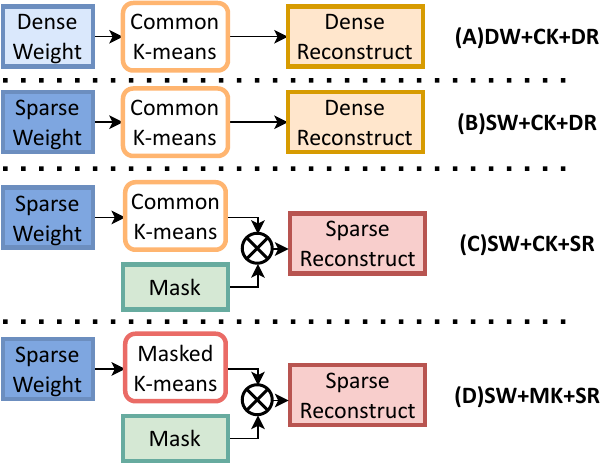}
    \caption{Illustration of the four comparing methods in ablation study. 'DW'/'SW' denotes Dense weight/Sparse weight to be vector clustered. 'CK'/'MK' denotes common k-means/masked k-means. 'DR'/'SR' denotes Dense reconstructed weight/Sparse reconstructed weight.}
    \label{fig:ablation study methods}
\end{figure}
A comprehensive ablation study is conducted on ResNet-18 to analyze and evaluate the impact of pruning on the vector clustering process. We compare four cases, which are shown in Fig.~\ref{fig:ablation study methods}. Case (A) represents the simplest vector quantization procedure. For the first two cases, (A) and (B), the mask is not involved in reconstructing compressed weights, so it doesn't need to be stored and there is no reduction in FLOPs. The difference between them is that sparse weights are clustered in (B). Storing the mask is required for cases (C) and (D) to reconstruct sparse weights and reduce FLOPs. Our method, represented by (D), uses masked k-means for vector clustering, which distinguishes it from the other three methods.
\begin{table}[h]
    \small
    \setlength{\tabcolsep}{2mm}
    \renewcommand{\arraystretch}{0.6}
    \centering
    \begin{tabular}{c|c|c|c}
    \toprule
    Cases & Total/Mask SSE & Flops & Acc (FP:69.7)\\
    \midrule
    A & 1153/463&  1.81G & 66.5\\
    B & 518/498 &  1.81G & 67.3\\
    C & 1840/1840& 0.54G & 61.1\\
    Ours & 251/251 & 0.54G\textcolor{blue}{(-70\%)} & 68.8 \textcolor{red}{(+1.5)}\\
    \bottomrule
    \end{tabular}
    \caption{Ablation study on ResNet-18: We ensure the same compression ratio of $\sim$22x for 4 cases.}
    \label{ablation stuty}
\end{table}

To reconstruct sparse weights, the bitmask needs to be stored. To ensure a fair comparison in terms of compression ratio, we set $k=1024$ (Resp. $k=512$) and $d=8$ (Resp. $d=16$) for Case A\&B (Resp. C\&D). We use Total SSE to indicate the cluster error of all weights and Mask SSE to indicate the cluster error of masked weights. When comparing the results of (A) and (B) in Tab.\ref{ablation stuty}, it is observed that applying sparse weights significantly reduces the total clustering error, but the clustering error for important weights remains large. This indicates that important weights are still forced to be aligned to many zeros. Our proposed method (D), which combines pruning and masked k-means, reduces the clustering error by 85\% compared to case (C). This suggests that the codewords can better approximate the important weights by using masked k-means.

Next, the accuracy and FLOPs of the four methods are evaluated. The results are tabulated in Tab.~\ref{ablation stuty}. The results of case (C) show that the conventional k-means clustering method is ineffective in reconstructing sparse weights with a restricted number of centroids. Our method improves accuracy and reduces FLOPs compared to (A) and (B), even with fewer codewords. This indicates that the cost of storing masks is worthwhile. 
\begin{figure*}
    \centering
    \includegraphics[width=0.9\linewidth]{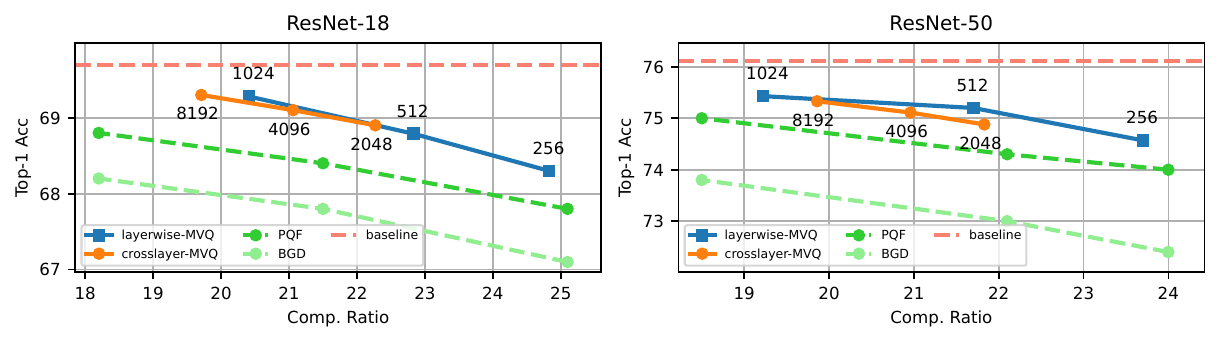}
    \caption{Comparsion with previous VQ-based methods on ResNet-18 and ResNet-50.}
    \label{fig:codebook_compare}
\end{figure*}
\subsection{Comparison with other methods}
In this subsection, We compare exclusively on ResNets and MaskRCNN with previous VQ-based methods, including BGD~\cite{stock2019and}, PQF~\cite{Martinez_2021_CVPR}. For other models not compressed by PQF, we compare with PvQ~\cite{kuzmin2023pruning} which conducts uniform scalar quantization on a larger set of models.
\begin{table}[t]
\small
    \centering
    \renewcommand{\arraystretch}{0.7}
    \setlength{\tabcolsep}{1mm}
    \begin{tabular}{lcllcl}
    \toprule
    Models & CR & Methods & Acc & Sparsity & FLOPs\\
    \midrule
    \multirow{3}{*}{\shortstack{SSL-\\ResNet-50\\Acc: 79.2}} & \multirow{3}{*}{$\sim$22$\times$} & MVQ(Ours) &  \textbf{{77.5}} & \textbf{75}\% & \textbf{1.11}G\\
    & & PQF~\cite{Martinez_2021_CVPR} &  77.1 & 0\% & 4.09G\\
    & & BGD~\cite{stock2019and} &  76.1 & 0\% & 4.09G\\
    \midrule

    \multirow{2}{*}{\shortstack{MobileNetV1\\Acc: 70.8}} & $\sim$17$\times$ & \multirow{2}{*}{MVQ(Ours)} &  \textbf{{66.3}} & \textbf{50}\% & \textbf{0.29}G\\
    & $\sim$19$\times$& &  64.3 & 50\% & 0.56G\\
    \midrule

    \multirow{2}{*}{\shortstack{MobileNetV2\\Acc: 71.7}} & \multirow{2}{*}{$\sim$16$\times$} & MVQ(Ours) &  \textbf{{65.1}} & \textbf{50}\% & \textbf{0.15}G\\
    & & PvQ~\cite{kuzmin2023pruning} &  59.1 & 0\% & 0.30G\\
    \midrule

    \multirow{2}{*}{\shortstack{EfficientNet\\Acc: 75.4}} & \multirow{2}{*}{$\sim$16$\times$} & MVQ(Ours) &  \textbf{{68.2}} & \textbf{50}\% & \textbf{0.14}G\\
    & & PvQ~\cite{kuzmin2023pruning} &  60.9 & 0\% & 0.28G\\
    \midrule

    \multirow{2}{*}{\shortstack{AlexNet\\Acc: 56.5}} & \multirow{2}{*}{$\sim$25$\times$} & \multirow{2}{*}{MVQ(Ours)} &  \multirow{2}{*}{\textbf{55.4}} & \multirow{2}{*}{\textbf{75}\%} & \multirow{2}{*}{\textbf{0.19}G}\\
    &  &  &   &  & \\
    \midrule

    \multirow{2}{*}{\shortstack{VGG-16\\Acc: 71.7}} & $\sim$28$\times$ & \multirow{2}{*}{MVQ(Ours)} &  \textbf{{69.7}} & \textbf{81}\% & \textbf{2.90}G\\
    & $\sim$34$\times$ & &  \textbf{{68.8}} & \textbf{88}\% & \textbf{1.96}G\\
    
    \bottomrule
    \end{tabular}
    \caption{Comparsion with other methods on more models. 'CR' denotes Compression Ratio.}
    \label{table:compare with diferent work}
\end{table}

\textbf{Image Classification.} Accuracy and compression ratio results on ImageNet are presented in Fig.~\ref{fig:codebook_compare} and Tab.~\ref{table:compare with diferent work}. We conduct experiments on both layerwise clustering (i.e. using one codebook for each layer) and crosslayer clustering(i.e. employing one codebook for all convolutional layers). Our results indicate that layerwise clustering yields a superior memory-accuracy trade-off. Our method outperforms all counterparts on a diverse set of models. For example, on the ResNet-18, our method delivers an accuracy improvement of 0.5\% to 0.8\% over PQF at compression ratios between 20× and 25×. Similarly, for the ResNet-50, our method reaches 75.2\% accuracy at a $\sim$22$\times$ compression ratio. This represents a 1.0\% improvement over the state-of-the-art. Furthermore, On MobileNets and EfficientNet, our approach demonstrates superior results compared to 2-bit uniform quantization. In addition to the increase in accuracy, our method also significantly reduces FLOPs by about 70\%(50\% for parameter-efficient models).

\begin{table}[t]
    \centering
    \small
    \setlength{\tabcolsep}{3mm}
    \renewcommand{\arraystretch}{0.6}
    \begin{tabular}{c|cc|cc}
    \toprule
    \multirow{2}{*}{Methods} & \multicolumn{2}{c|}{ResNet-18} & \multicolumn{2}{c}{ResNet-50}\\
    \cmidrule{2-5}
    & SSE &  Acc & SSE & Acc\\
    \midrule
    PQF~\cite{Martinez_2021_CVPR} & 605 & 68.2 & 1150 & 74.2\\
    \midrule
    Ours & \textbf{251} & \textbf{68.8} & \textbf{336} & \textbf{75.2}\\
    \bottomrule
    \end{tabular}
    \caption{Comparison of the SSE and accuracy between our method and previous vector quantization approach at $\sim 22\times$ compression ratio. 
    SSE is evaluated before fine-tuning.}
    \label{tab:sse_with_PQF}
\end{table}

\begin{table}[t]
    \small
    \setlength{\tabcolsep}{2mm}
    \renewcommand{\arraystretch}{0.6}
    \centering
    \begin{tabular}{c|c|c|c|c|c}
    \toprule
    \multicolumn{6}{c}{MaskRCNN on COCO2017 dataset} \\
    Methods & CR & Sparsity & FLOPs & $AP^{bb}$ & $AP^{mk}$ \\
     \midrule
    Baseline & - & 0\% & 88.9G & 37.9 & 34.6 \\
    BGD~\cite{stock2019and} & 26$\times$ & 0\% & 88.9G & 33.9 &30.8 \\
    PQF~\cite{Martinez_2021_CVPR} & 26$\times$ & 0\% & 88.9G & 36.3 & 33.5\\
    Ours & 26$\times$ & \textbf{75}\% & \textbf{23.4}G& \textbf{36.8}  & \textbf{33.8}\\ 
    \midrule
    \multicolumn{6}{c}{DeepLab-V3 on VOC dataset} \\
    Methods & CR & Sparsity & FLOPs & \multicolumn{2}{c}{mIoU} \\
    \midrule
    Baseline & - & 0\% & 4.2G & \multicolumn{2}{c}{72.9} \\
    PvQ~\cite{kuzmin2023pruning} & 16$\times$ & 0\% & 4.2G &\multicolumn{2}{c}{17.6}\\
    Ours & 19$\times$ & \textbf{50}\% & \textbf{2.1}G & \multicolumn{2}{c}{\textbf{66.5}}\\ 
    \bottomrule
    \end{tabular}
    \caption{Comparing bounding box and mask metrics on COCO 2017 with other vector quantization methods. FLOPs are calculated with an input size of [3,800,800] for COCO and [3,512,512] for VOC }
    \label{tab:maskrcnn results}
\end{table}

To clarify how our method achieves greater accuracy than other weight clustering methods with the same compression ratio, we conduct a comparison of SSE between our work and PQF. The results are shown in Tab.~\ref{tab:sse_with_PQF}. At the same compression ratio, our method reaches a significantly lower SSE than PQF, especially on ResNet-50, which results in smaller accuracy degradation.

\textbf{Object Detection and Segmentation.} Our method is also validated on the task of object detection and segmentation. We compress the ResNet-50 Mask-RCNN FPN architecture using the MS COCO 2017 dataset~\cite{lin2014microsoft}. The results are represented in Tab.~\ref{tab:maskrcnn results}. Our method achieves a box AP of 36.8, and a mask AP of 33.8 at 26$\times$ compression ratio. Additionally, our method reduces the FLOPs by 70\% compared with previous vector quantization methods. We also compress DeepLab-v3~\cite{chen2017rethinking} with MobileNet-V2 backbone trained for semantic segmentation on Pascal VOC~\cite{everingham2010pascal}. Our MVQ maintains a high mIoU at 19$\times$ compression ratio while 2bit uniform quantization crashes.

\section{Hardware Implementation Experiments and Performance Evaluation}
\subsection{System setup}
We consider the open-source PULP~\cite{pullini2019mr} framework to validate the performance of our accelerator. Our system consists of a main control System-on-Chip (SoC) and an accelerator subsystem built around a customized systolic array used for DNN inference acceleration. The detailed mechanism of SoC and hierarchy caches has been discussed in ~\cite{wang2024ews}

To assess the inference performance of our accelerator, we conducted tests by running five CNN models (ResNet-18, ResNet-50, VGG-16, AlexNet, MobileNet-v1) on three scales of PE arrays (16$\times$16, 32$\times$32, and 64$\times$64). Throughout, we compare designs in terms of area efficiency, speedup, and energy efficiency. Our ASIC design was synthesized using the Synopsys Design Compiler with a 40 nm 0.99V LVT Library. The accelerator's performance was assessed through gate-level netlist simulation, while power consumption during operation was evaluated using Synopsys Prime Time PX (PT).

For a thorough comparison, we conducted tests using 6 different hardware settings based on WS and EWS dataflow:

(a) WS(WS-base): This is the baseline setting for WS dataflow where weights are simply quantized to 8-bit.

(b) WS-CMS: This setting fully utilizes MVQ within WS dataflow.

(c) EWS(EWS-base): This is the baseline setting where weights are simply quantized to 8-bit.

(d) EWS-C: This setting uses common vector quantization~\cite{stock2019and, Martinez_2021_CVPR} to compress weights.

(e) EWS-CM: This setting utilizes our proposed masked vector quantization to compress weights.

(f) EWS-CMS: This setting involves updating sparse tiles based on the EWS-CM approach, i.e., this setting fully utilizes MVQ within EWS dataflow

To ensure a fair comparison in terms of compression ratio, we set $k=1024$ and $d=8$ for EWS-C, while $k=512$ and $d=16$ for EWS-CM and EWS-CMS.

\subsection{Area Analysis}
\begin{table}[t]
    \centering
    \renewcommand{\arraystretch}{0.6}
    \small
    \begin{tabular}{c|c|c|c|c}
    \toprule
    \multicolumn{2}{c|}{} & Size-16 & Size-32 & Size-64\\
    \midrule
    \multirow{5}{*}{\makecell{Accle-\\rator}} & WS & 0.188 & 0.734 & 2.812\\
    \cmidrule{2-5}
    & EWS & 0.36 & 1.14 & 4.236\\
    \cmidrule{2-5}
    & EWS-C/CM & 0.650 & 1.505 & 4.776\\
    \cmidrule{2-5}
    & EWS-CMS & 0.469 & 0.828 & 2.129 \\
    \midrule
    \multicolumn{2}{c|}{L1} & 0.484 & 0.968 & 0.968\\
    \midrule
    \multicolumn{2}{c|}{L2} & \multicolumn{3}{c}{6.924} \\
    \midrule
    \multicolumn{2}{c|}{Others} & 0.787 & 1.303 & 1.659 \\
    \bottomrule
    \end{tabular}
    \caption{Area comparison on 3 array scales}
    \label{area}
\end{table}
The areas of the accelerator under different hardware settings at 75\% sparsity are summarized in Tab.~\ref{area}, which were obtained through DC synthesis. EWS-C/EWS-CM and EWS-CMS incur an additional area cost for the codebook reg file. As the array size increases, the area of the CRF also increases since it has to support more CRF read ports. However, our EWS-CMS, with its sparse tile implementation, significantly reduces the area cost of the systolic array by 50\%-60\%. Even with the area cost of CRF, our EWS-CMS for 64$\times$64 array reduces the area by 55\% compared to base EWS. Additionally, Tab.~\ref{area} includes the area information of L1, L2, and other system components. While L2 is fixed at 2MB, we tested different L1 sizes and found that for 16$\times$16 arrays (Resp. 32$\times$32 and 64$\times$64 arrays), the benefits are limited beyond 128KB (Resp. 256KB). Therefore, we set the L1 size to 128KB for the 16$\times$16 array (Resp. 256KB for the 32$\times$32 and 64$\times$64 arrays). Different areas are observed for other components, including DMA, peripheral interfaces, and system interconnections, under three array sizes.

\subsection{Energy Analysis}
\begin{table}[t]
    \small
    \renewcommand{\arraystretch}{0.6}
    \centering
    \begin{tabular}{c|c|c|c|c|c|c|c}
    \toprule
     & DRAM & L2 & L1 & PRF & ARF & WRF & CRF\\
     \midrule
     Energy & 200 & 15 & 6 & 0.22 & 0.11 & 0.02 & 0.02\\
     \bottomrule
    \end{tabular}
    \caption{Normalized data access energy cost for different levels of storage. We set the unit cost to a MAC operation.}
    \label{data-access-model}
\end{table}

\begin{figure}[t]
    \centering
    \includegraphics[width=1\linewidth]{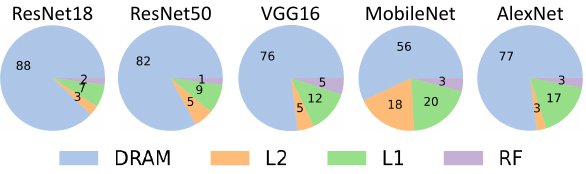}
    \caption{Data access cost ratio of different levels of memory}
    \label{access_pie}
\end{figure}
Firstly, we model the data access energy cost based on Tab.\ref{data-access-model}. The DRAM result is obtained from ~\cite{sim2019energy} and ~\cite{chen2016eyeriss}, while the other results are obtained by PT analysis. The data access cost of different levels of memory is presented in Fig.~\ref{access_pie}, from which it's clear that DRAM access overhead accounts for the majority. By applying our masked vector quantization technique, the storage cost of 8-bit weights can be greatly reduced, leading to a significant decrease in energy cost for DRAM access. 
\begin{figure}[h]
    \centering
    \includegraphics[width=1\linewidth]{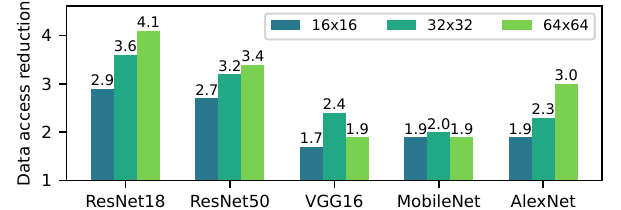}
    \caption{Data access cost reduction by employing MVQ compression}
    \label{data_access_reduction}
\end{figure}
The total data access cost reduction results are summarized in Fig.~\ref{data_access_reduction}. By employing MVQ. the access cost of ResNet18 can be reduced up to 4.1$\times$. For MobileNets, our approach also reduces the cost by 2$\times$. We have to mention that for VGG16, input features of the first few layers are too large that they have to be stored in DRAM, thus the reduction ratio is lower than expected.
\begin{figure*}[t]
    \centering
    \includegraphics[width=1\linewidth]{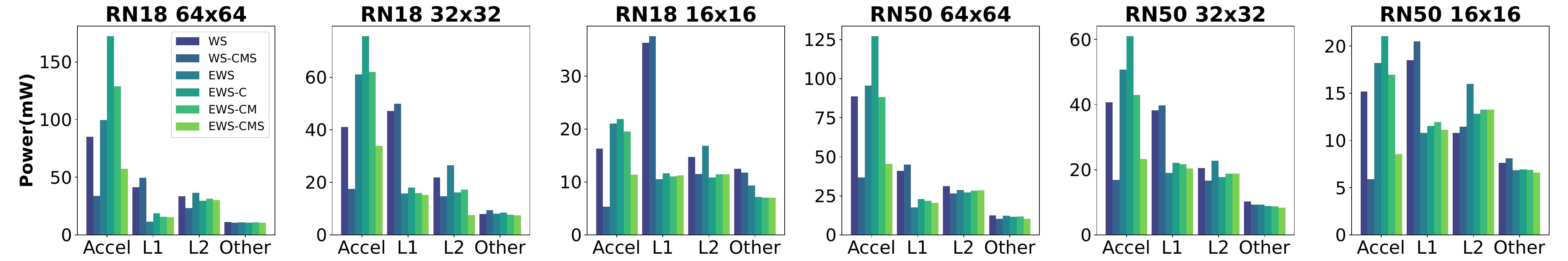}
    \caption{Power consumption breakdown for hardware under test. 'Accel' means the main accelerator which consists of the systolic array, controller and RF. 'Others' consists CPU, DMA, interface and IO.}
    \label{energy_breakdown}
\end{figure*}

Secondly, we present a detailed power breakdown of our accelerator. As depicted in Fig.\ref{energy_breakdown}. WS has a very high L1 access cost due to insufficient data reusability, which is also why EWS surpasses WS. Benefiting from the sparse tile design, our EWS-CMS/WS-CMS greatly reduces the energy cost of the main accelerator (PE Array), which accounts for the majority of total power consumption. As the array size increases, the effectiveness of EWS-CMS/WS-CMS becomes more pronounced.

\subsection{Performance Analysis}
\begin{figure}[h]
    \centering
    \includegraphics[width=0.95\linewidth]{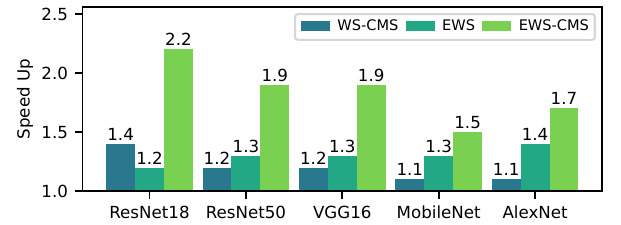}
    \caption{Speed up comparison of different hardware settings on various models. The array size is fixed to 64*64.}
    \label{speedup_all}
\end{figure}

\begin{figure}[h]
    \centering
    \includegraphics[width=0.95\linewidth]{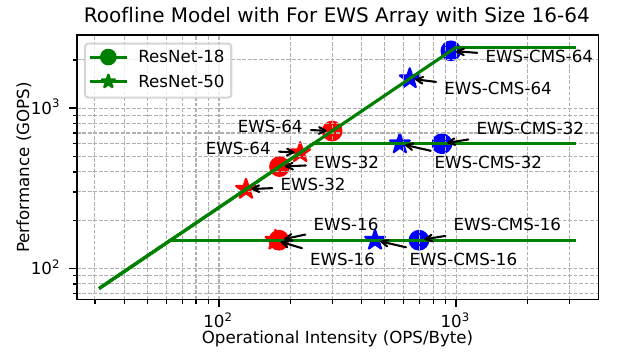}
    \caption{Roofline Model for EWS Array with different sizes}
    \label{roofline}
\end{figure}

Our accelerator also provides a modest speedup. Results are shown in Fig.~\ref{speedup_all}. Our EWS-CMS achieves a 1.2-1.8$\times$ speedup over EWS when the array size is set to 64$\times$64. We conducted an analysis using the roofline theory to understand the reason for this improvement. As depicted in Fig.~\ref{roofline}, the performance of the accelerator is limited by the weight-loading datawidth for array sizes larger than 32$\times$32. With the aid of our compression algorithm, the datawidth requirement for weight-loading is significantly reduced, as we now load only the index from L2 SRAM instead of the complete weights. As a result, the speedup effect of weight compression becomes more noticeable as the array size increases. The speed-up of WS-CMS is less significant because the frequent L1 access greatly constrains the performance of WS dataflow. 

\begin{table*}[t]
\setlength{\tabcolsep}{6pt}
{\fontsize{8.4}{9.7}\selectfont 
\begin{tabular}{c|c| c |c |c| c| c | c |c } 
         \hline
          & SparTen~\cite{gondimalla2019sparten} &  CGNet~\cite{hua2019boosting} &  SPOTS~\cite{SPOTS2022} & S2TA~\cite{liu2022s2ta} & \multicolumn{4}{c}{\textbf{Ours}}  \\ 
          & MICRO19 & MICRO19 & TACO22 & HPCA22 & MVQ-16 & MVQ-32 & \multicolumn{2}{c}{MVQ-64} \\ \hline
         Process(nm) & 45 & 28 &  45 & 16/65 &  \multicolumn{4}{c}{40} \\ \hline
         Freq(GHz) & 0.8 & 0.5(0.9V)  & 0.5 & 1/0.5 & \multicolumn{4}{c}{0.3} \\ \hline 
         SRAM Size & NA & 606K+576K  & 1M+512K & 2M+512K &  2M+128KB& \multicolumn{3}{c}{2M+256K} \\ \hline
         MACs & 32 & 576  & 512 & 2048  & 64 & 256 & 1024 &  1024(2048)\\ \hline
         Sparse Granularity & Random & Channel-wise  & Group-wise & N:M  & \multicolumn{4}{c}{N:M} \\ \hline
         Sparsity & NA & 60\% 	& 27\% & 50\% & 75\% & 75\%  & 75\% & 75\%(50\%) \\ \hline
         Quantization & INT8 & INT8  & INT16 & INT8  & INT8 &INT8 & INT8 &INT8\\ \hline
         Compression Ratio & NA & 10$\times$ & 3$\times$ & 6.4$\times$ &22$\times$ & 22$\times$ & 22$\times$ & 25$\times$ \\ \hline
         Workload & AlexNet & ResNet18  & VGG16 & AlexNet & ResNet18 & ResNet18 & ResNet18 & AlexNet\\ \hline
         Dataflow & OS & WS & OS  & OS  & \multicolumn{4}{c}{EWS} \\ \hline
         Peak Performance(TOPS) & 0.2 & 2.4  & 0.5 & 8/4  &0.15 & 0.6 & 2.4 & 2.4(4.8) \\ \hline
         Area(mm$^2$) & 0.766 & 5.574  & 8.61 & 3.8/24 & 8.66 & 10.02 & 11.68 & 11.68(12.48)\\ \hline
         Efficiency(TOPS/W) & 0.68 & 4.5  & 0.47 & 14/1.1 &2.3 & 4.1 & 6.9 & 4.4(3.8)
         \\ \hline
         N-Efficiency(TOPS/W) & 0.97 & 2.43  & 0.67 & 1.64/2.19 &2.3 & 4.1 & 6.9 & 4.4(3.8)\\ \hline
\end{tabular}
\caption{Comparison with Other Works. We refer to ~\cite{stillmaker2017scaling} to normalize energy efficiency into 40nm process.}
\label{acc_comp}
}
\end{table*}

\subsection{Efficiency Analysis}

\begin{figure}[h]
    \centering
    \includegraphics[width=0.95\linewidth]{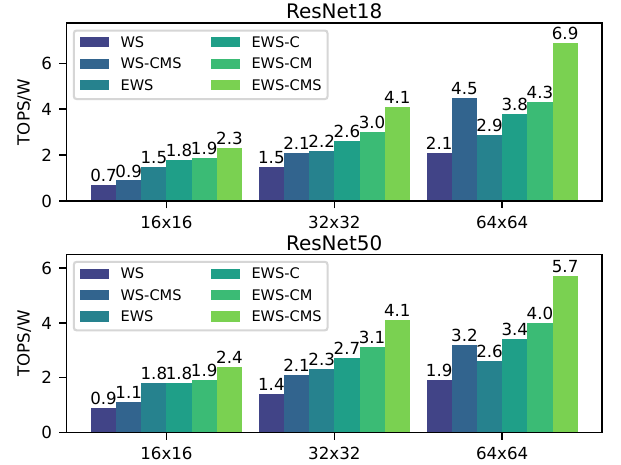}
    \caption{Energy efficiency comparison of different hardware settings on three array sizes}
    \label{TOPS_W}
\end{figure}

\begin{figure}[h]
    \centering
    \includegraphics[width=0.95\linewidth]{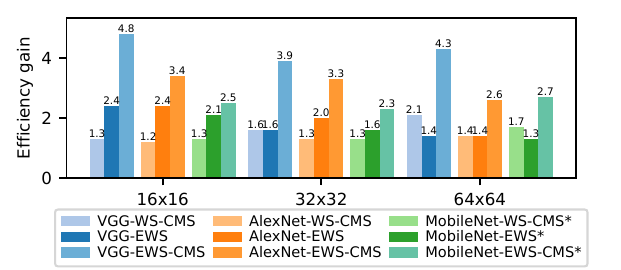}
    \caption{Energy efficiency gain compared to WS baseline on more models. * denotes pointwise convolution only.}
    \label{TOPS_W2}
\end{figure}

Then, we compared the energy efficiency of five accelerators (measured in TOPS/W) using different array sizes. A detailed comparison on ResNets can be found in Fig.~\ref{TOPS_W}. It's worth noting that the energy consumption shown in Fig.~\ref{TOPS_W} does not include main memory access, which typically accounts for a significant portion of the total energy consumption. Our proposed accelerator benefits from both reduced energy consumption and enhanced performance, showing significant efficiency improvements for both WS and EWS dataflows. Generally, as the array size increases, energy efficiency tends to improve. In comparison to the EWS baseline, our proposed EWS-CMS accelerator achieves 53.3\% to 137.9\% higher energy efficiency on ResNet18. Likewise, our WS-CMS accelerator achieves 28.5\% to 114.2\% higher energy efficiency relative to the WS baseline. It is noteworthy that the L1 power consumption is higher in the WS dataflow, which leads to a slightly smaller MVQ improvement for WS compared to EWS.

Furthermore, the efficiency improvements for VGG16, AlexNet, and MobileNetV1 are depicted in Fig.~\ref{TOPS_W2}. Our proposed accelerator demonstrates an average efficiency improvement 46\% and 90\% for WS and EWS dataflow, respectively. It is important to note that for WS/EWS dataflows, depthwise convolution requires mapping weight elements to the diagonal of the PE array. This results in the low performance of the depthwise layer as only the diagonal elements are involved in the computation. Moreover, the number of parameters in the depthwise layer is quite limited, meaning that the hardware operation bottleneck does not stem from the weight-loading process. As a result, depthwise layers don't fully utilize the advantages brought by MVQ. Therefore, we present pointwise convolution results only. 

\subsection{Comparision with other sparse CNN accelerators}

Our MVQ is compared with a series of sparse CNN accelerators, including SparTen~\cite{gondimalla2019sparten}, CGNet~\cite{hua2019boosting}, SPOTS~\cite{SPOTS2022}, and S2TA~\cite{liu2022s2ta}.
Additionally, we have normalized the energy efficiency for a 40 nm process to ensure a fair comparison. Compared to existing works, our design greatly reduces the data access cost with MVQ which achieves a very high compression ratio.  
As depicted in Tab.~\ref{acc_comp}, compared to S2TA~\cite{liu2022s2ta} which reports the highest efficiency and the lowest power consumption, our MVQ achieves 73\% higher energy efficiency on AlexNet. This highlights the efficiency of our design.

\section{Conclusion}
This paper proposes a highly efficient accelerator with a Masked Vector Quantization (MVQ) compression method. At the algorithm level, to address the issue that important weights are often forced to align with unimportant ones, N:M pruning has been performed within each subvector to remove the unimportant weights. Subsequently, the mask k-means algorithm has been applied to prevent pruned weights from interfering with the vector clustering process. As a result, our method significantly reduces the vector clustering error compared with existing approaches. At the architecture level, we implement an area- and energy-efficient accelerator based on the EWS dataflow. Employing an index-based weight-loading unit and a sparsity-aware systolic array, our accelerator fully realizes the benefits brought by MVQ. 

Algorithm experiments on various models demonstrate that our approach reaches higher accuracy at the same compression ratios and reduces FLOPs. In 40nm ASIC post-synthesis evaluation, our proposed MVQ-based accelerator can boost energy efficiency by 2.3$\times$ and cut the array size by 55\% compared to the base EWS accelerator. This has also led to a superior performance of our accelerator against prior arts under energy efficiency (1.73$\times$).

\begin{acks}
 We thank our shepherd, Macro Donato, for his ongoing support and guidance during the revision process. We also thank the anonymous reviewers for their constructive feedback for improving the work.
This work was supported by the National Natural Science Foundation of China under Grant 62274142 and the Shenzhen Science and Technology Program under Grant KJZD20230923115213027.
\end{acks}

\bibliographystyle{plain}
\bibliography{main}

\end{document}